\documentclass{article}

\usepackage{microtype}
\usepackage{graphicx}
\usepackage{booktabs}
\usepackage{hyperref}
\usepackage[accepted]{icml2024}

\usepackage{amsmath}
\usepackage{amssymb}
\usepackage{mathtools}
\usepackage{amsthm}
\usepackage[capitalize,noabbrev]{cleveref}
\usepackage{xspace}
\usepackage{subcaption}

\theoremstyle{plain}

\theoremstyle{definition}

\theoremstyle{remark}

\usepackage[textsize=tiny]{todonotes}

\def\ie{\emph{i.e.}\@\xspace}
\def\vs{\emph{vs.}\@\xspace}

\newcommand{\app}{\raise.17ex\hbox{$\scriptstyle\sim$}}

\begin{document}
\twocolumn[
\icmltitle{Humanoid Locomotion as Next Token Prediction}
\begin{icmlauthorlist}
\icmlauthor{Ilija Radosavovic}{yyy}
\icmlauthor{Bike Zhang}{yyy}
\icmlauthor{Baifeng Shi}{yyy}
\icmlauthor{Jathushan Rajasegaran}{yyy}
\icmlauthor{Sarthak Kamat}{yyy}\\
\icmlauthor{Trevor Darrell}{yyy}
\icmlauthor{Koushil Sreenath}{yyy}
\icmlauthor{Jitendra Malik}{yyy}
\end{icmlauthorlist}
\icmlaffiliation{yyy}{University of California, Berkeley}
\icmlcorrespondingauthor{Ilija Radosavovic}{ilija@berkeley.edu}
\icmlkeywords{Machine Learning, ICML}
\vskip 0.3in
]
\printAffiliationsAndNotice{}

\begin{abstract}
We cast real-world humanoid control as a next token prediction problem, akin to predicting the next word in language. Our model is a causal transformer trained via autoregressive prediction of sensorimotor trajectories. To account for the multi-modal nature of the data, we perform prediction in a modality-aligned way, and for each input token predict the next token from the same modality. This general formulation enables us to leverage data with missing modalities, like video trajectories without actions. We train our model on a collection of simulated trajectories coming from prior neural network policies, model-based controllers, motion capture data, and YouTube videos of humans. We show that our model enables a full-sized humanoid to walk in San Francisco zero-shot. Our model can transfer to the real world even when trained on only 27 hours of walking data, and can generalize to commands not seen during training like walking backward. These findings suggest a promising path toward learning challenging real-world control tasks by generative modeling of sensorimotor trajectories.
\end{abstract}
\section{Introduction}

The last decade of artificial intelligence (AI) has shown that large neural networks trained on diverse datasets from the Internet can lead to impressive results across different settings. The core enablers of this wave of AI have been large transformer models~\cite{Vaswani2017} trained by generative modeling of massive quantities of language data from the Internet~\cite{Radford2018, Devlin2019, Radford2019, Radford2021, Brown2020}. By predicting the next word, these models acquire rich representations of language that can be transferred to downstream tasks~\cite{Radford2018}, perform multi-task learning~\cite{Radford2019, Radford2021}, and learn in a few-shot manner~\cite{Brown2020}.

Are such modeling techniques exclusive to language? Can we learn powerful models of sensory and motor representations in a similar fashion? Indeed, we have seen that we can learn good representations of high-dimensional visual data by autoregressive modeling~\cite{Chen2020generative} and related masked modeling approaches~\cite{He2021}. While there has been positive signal on learning sensorimotor representations in the context of manipulation~\cite{radosavovic2023robot}, this area remains largely unexplored.

In this paper, we cast humanoid control as data modeling of large collections of sensorimotor trajectories. Like in language, we train a general transformer model to autoregressively predict shifted input sequences. In contrast to language, the nature of data in robotics is different. It is high-dimensional and contains multiple input modalities. Different modalities include sensors, like joint encoders or inertial measurement units, as well as motor commands. These give rise to \emph{sensorimotor trajectories} which we view as the \emph{sentences} of the physical world. Adopting this perspective suggests a simple instantiation of the language modeling framework in the robotic context. We tokenize the input trajectories and train a causal transformer model to predict shifted tokens. Importantly, we predict \emph{complete} input sequences, including both sensory and motor tokens. In other words, we are modeling the \emph{joint} data distribution as opposed to the conditional action distribution.

This has several benefits. First, we are training the neural network to predict more bits of information and consequently acquire a richer model of the world. Second, we can leverage noisy or imperfect trajectories that may contain suboptimal actions. Third, we can generalize our framework to learning from trajectories with missing information.

\begin{figure*}[t!]\centering\vspace{0mm}
\includegraphics[width=1.0\linewidth]{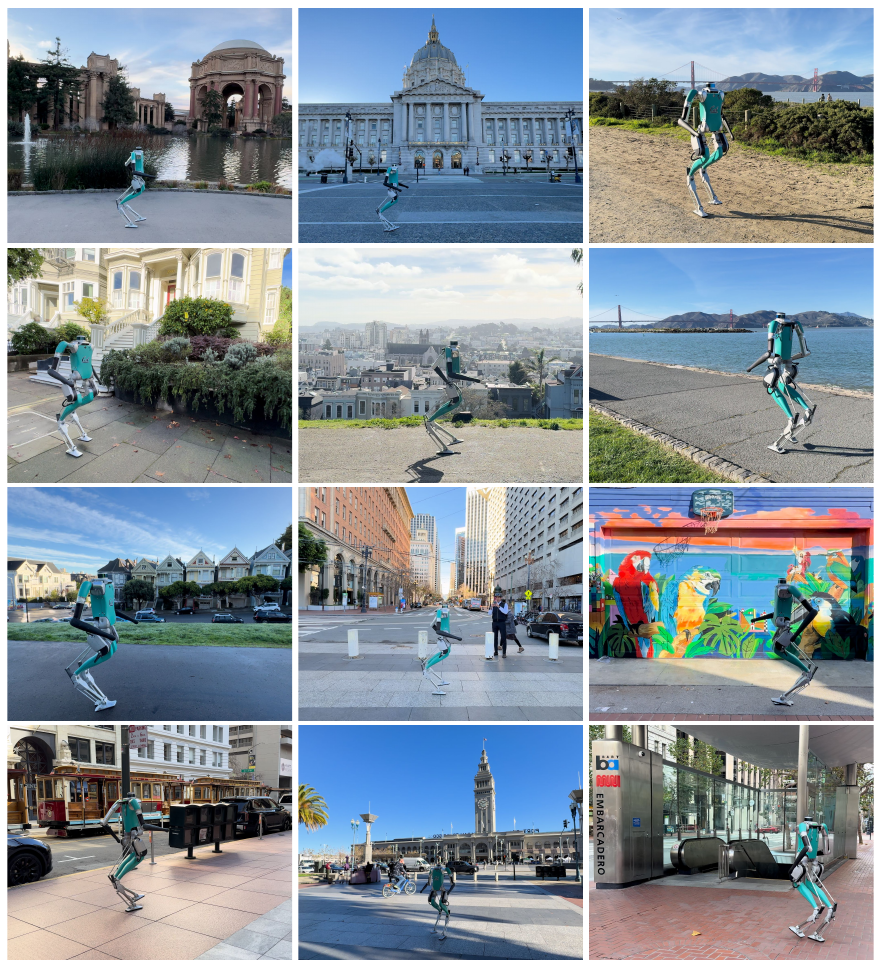}\vspace{0mm}
\caption{\textbf{A humanoid that walks in San Francisco.} We deploy our policy to various locations in San Francisco over the course of one week. Please see our~\href{humanoid-next-token-prediction.github.io}{project page} for videos. We show that our policy can walk over different surfaces including walkways, concrete, asphalt, tiled plazas, and sanded roads. We find that our policy follows omnidirectional velocity commands well and enables deployment in a challenging city environment like San Francisco.\\ \\ \\}
\label{fig:sf}\vspace{0mm}
\end{figure*}

Our core observation is that if a trajectory is incomplete, \ie, some of the sensory or motor information is missing, we can still learn from it by predicting whatever information is present and replacing the missing tokens with learnable mask tokens. The intuition is that if the model has learned to make good predictions, even in the absence of information, it will have acquired a better model of the physical world. A very important source of such data are human videos from the Internet. Namely, we can observe human movement in videos but we do not get access to the motor commands or complete sensory inputs. We demonstrate that our method can learn from such data sources effectively.

\newpage

To validate our method, we apply it to the challenging task of real-world humanoid locomotion. We use the full-sized Digit humanoid robot developed by Agility Robotics. We first collect a dataset of sensorimotor trajectories in simulation. These include complete trajectories from a neural network policy trained with reinforcement learning~\cite{RealHumanoid2023}, as well as incomplete trajectories from three different sources: (i) Agility Robotics controller based on model predictive control, (ii) motion capture of humans, and (iii) YouTube videos of humans. We reconstruct human videos by using computer vision techniques and retarget both motion capture and YouTube trajectories via inverse kinematics. We then train a transformer model to autoregressively predict trajectories. At test time, we execute the actions autoregressively and ignore the sensory predictions.

We demonstrate that our policy can be deployed in the real world zero-shot and walk on different surfaces. Specifically, deploy our model across a range of different locations in San Francisco over the course of one week. Please see Figure~\ref{fig:sf} for examples and our \href{humanoid-next-token-prediction.github.io}{project page} for videos. To quantitatively evaluate different aspects of our approach, we perform an extensive study in simulation. We find that our autoregressive policies trained from offline data alone are comparable to the state-of-the-art approaches that use reinforcement learning~\cite{RealHumanoid2023} in tested settings. We further find that our approach can readily benefit from incomplete trajectories and has favorable scaling properties.

These findings suggest a promising path toward learning challenging real-world robot control tasks by generative modeling of large collections of  sensorimotor trajectories.

\begin{figure*}[t]\centering\vspace{0mm}
\includegraphics[width=1.0\linewidth]{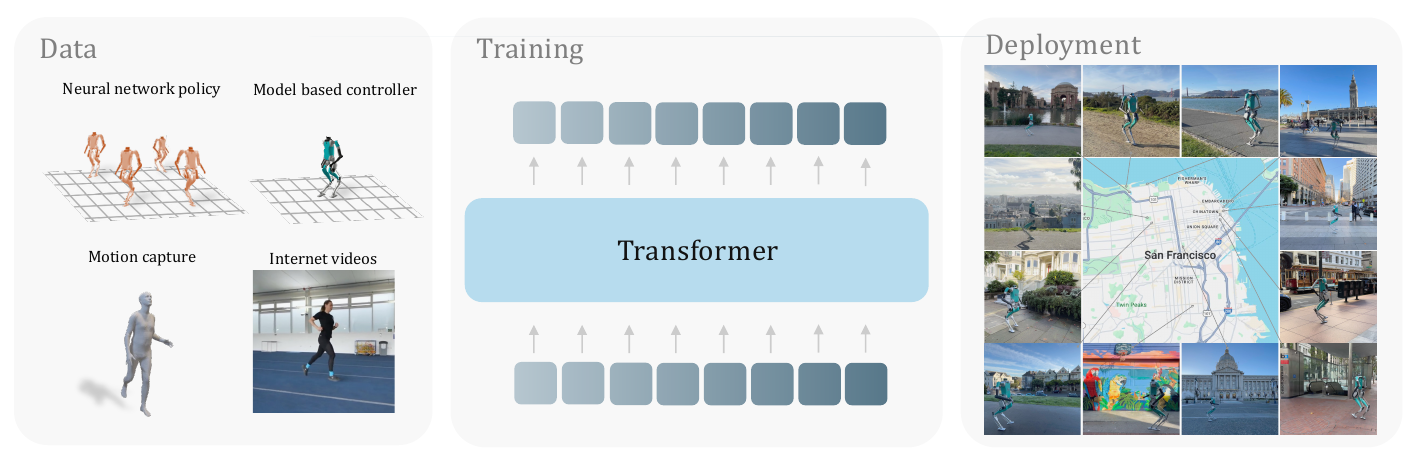}\vspace{-4mm}
\caption{\textbf{Humanoid locomotion as next token prediction.} We collect a dataset on trajectories from various sources, such as from neural network policies, model-based controllers, human motion capture, and YouTube videos of humans. Then we use this dataset to train a transformer policy by autoregressive modeling of observations and actions. Our transformer allows a humanoid to walk zero-shot on various terrains around San Francisco. Please see our \href{humanoid-next-token-prediction.github.io}{project page} for video results.}
\label{fig:overview}\vspace{0mm}
\end{figure*}

\section{Related Work}

\textbf{Generative modeling.} 
The study of data has been extensive, ranging from Shannon's foundational work~\cite{shannon1951prediction} to the recent era of large language models.
Various such models emerged over the last decade. Notable such models includes, GAN~\cite{goodfellow2014generative} and Diffusion models~\cite{sohl2015deep,ho2020denoising}
for generating pixels, LSTM~\cite{hochreiter1997long} and GPT~\cite{Radford2018} for generating language tokens. These models have been adopted for other modalities as well~\cite{oord2016wavenet, engel2019gansynth, wu2016learning}. Among these, autoregressive transformer models became the front runner, due to the impressive scaling behaviours~\cite{kaplan2020scaling} and ability to learn from in-context examples~\cite{brown2020language}. This behavior is even shown to extend to other modalities such as pixels~\cite{Chen2020generative}, language-pixels~\cite{ramesh2021zero}, and language-pixels-audio~\cite{kondratyuk2023videopoet}. We explore autoregressive generative models in the context of real-world humanoid locomotion. 

\textbf{Transformers in robotics.} Following the success of transformer models~\cite{Vaswani2017} in natural language processing~\cite{Radford2018,Devlin2019,Radford2019,brown2020language} and computer vision~\cite{Dosovitskiy2020,He2021}, over the last few years, there has been an increased interested in using transformer models in robotics. We have seen several works showing that transformers can be effective with behavior cloning. For example, \cite{shridhar2022peract} learns multi-task transformer policies with language, and~\cite{brohan2022rt} trains language-conditioned manipulation policies from large-scale data. \cite{Driess2023} trains language models with embodied data. We have also seen that transformer policies can be effective for large-scale reinforcement learning~\cite{RealHumanoid2023}. \cite{radosavovic2023robot} learns sensorimotor representations with masked prediction. \cite{bousmalis2023robocat} trains goal-conditioned policies are learned from demonstrations. Likewise, we share the goal of using transformer models for robotics but focus on autoregressive modeling of diverse trajectories for real-world humanoid locomotion.

\textbf{Humanoid locomotion.} Mastering the ability for robots to walk has been a long-standing challenge in robotics. In the past several decades, roboticists have built a variety of humanoid robots \cite{Kato1973, Hirai1998, Nelson2012, Stasse2017, Chignoli2021} to explore human-like locomotion skills. Stable locomotion behaviors have been achieved through model-based control approaches \cite{Raibert1986, Kajita2001}, and optimization-based methods further enable highly dynamic humanoid motions \cite{Kuindersma2020}. Although significant progress has been made with these strategies and combining them with learning~\cite{castillo2021robust}, learning-based approaches are gaining attention for their ability to improve and adapt to a wide range of environments. Recently, we have seen that a purely learning based approach trained with large-scale reinforcement learning in simulation can enable real-world humanoid locomotion~\cite{RealHumanoid2023}. Like in prior work, our model is a causal transformer. Unlike prior work, we perform autoregressive modeling instead of reinforcement learning.

\section{Approach}\label{sec:approach}

In this section, we assume that a dataset $\mathcal{D}$ of sensorimotor trajectories $\mathcal{T}$ is given and describe our approach below.

\subsection{Objective}

Each sensorimotor trajectory is a sequence of sensory observations and actions: $\mathcal{T} = (o_1, a_1, o_2, a_2, ..., o_T, a_T)$. We first tokenize the trajectory into K tokens to obtain $t = (t_1, t_2, t_3, ..., t_K)$. Our goal is to train a neural network to model the density function $p(t)$ autoregressively:
\begin{equation}
    p(t) = \prod_{k=1}^{K} p(t_k | t_{k-1}, ..., t_{1})
\end{equation}
We train our model by minimizing the negative log-likelihood over our trajectory dataset:
\begin{equation}
    L = \sum_{t \in \mathcal{D}} - \log p(t)
\end{equation}

\newpage

We assume a Gaussian distribution with constant variance and train a neural network to minimize the mean squared error between the predicted and the ground truth tokens:
\begin{equation}
    L = \frac{1}{K} \sum_{k = 1}^{K} (\widehat{t}_k - t_k)^2
    \label{eq:loss}
\end{equation}
Instead of regressing the raw token values, we could quantizing each dimension into bins or perform vector quantization. However, we found the regression approach to work reasonably well in practice and opt for it for simplicity.

\subsection{Missing modalities}

In the discussion so far we have assumed that each trajectory is a sequence of observations and actions. Next, we show how our framework can be generalized to sequences with missing modalities, like trajectories extracted from human videos that do not have actions. Suppose we are given a trajectory of observations without the actions $\mathcal{T} = (o_1, o_2, ..., o_T)$. Our key insight is that we can treat a trajectory without actions like a regular trajectory with actions masked. Namely, we can insert mask tokens $\texttt{[M]}$ to obtain $\mathcal{T} = (o_1, \texttt{[M]}, o_2, \texttt{[M]}, ..., o_T, \texttt{[M]})$. This trajectory now has the same format as our regular trajectories and thus can be processed in a unified way. We ignore the loss for the predictions that correspond to the masked part of inputs. Note that this principle is not limited to actions and applies to any other modality as well.

\subsection{Aligned prediction}

\begin{figure*}
\centering
\includegraphics[width=0.9\linewidth]{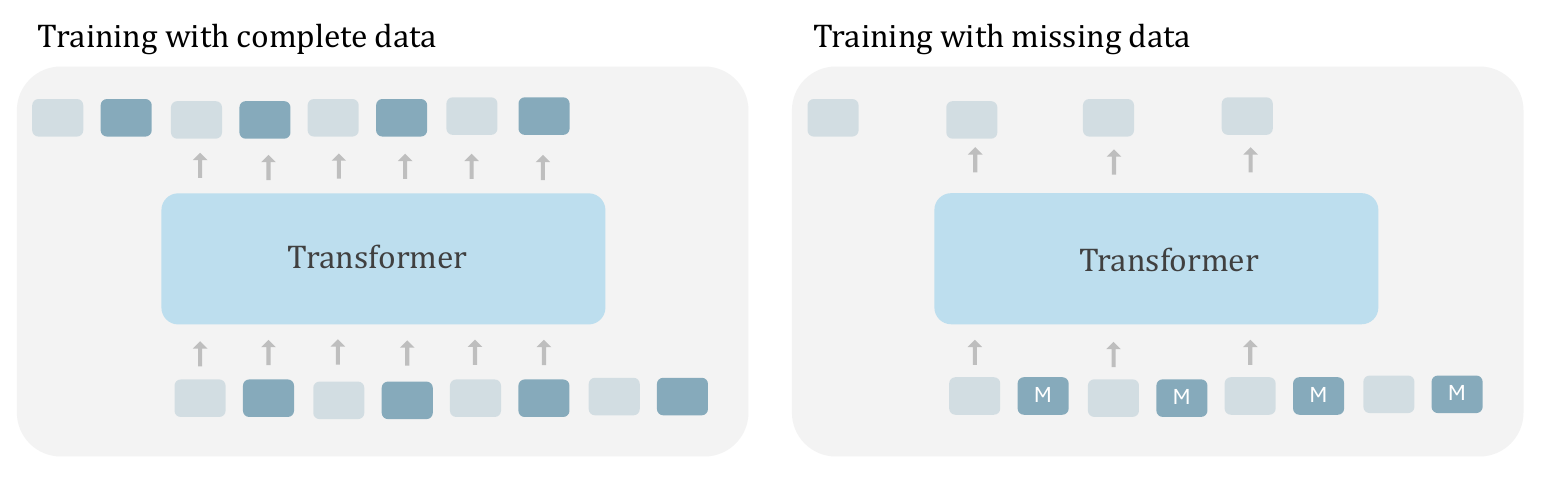}\vspace{-2mm}
\caption{\textbf{A general framework for training with different data sources.} Our data modeling allows us to train our transformer with multiple modes of training. In the case of observation-action pairs being available, we train our transformer to predict the next pair of observation-action. When there is no action data available, with MoCap and internet data, we only train our transformer to predict the next observations by masking the actions with a mask token. These two models of training allow our model to utilize both types of data, and this enables us to scale our training in terms of data. }
\label{fig:arch}\vspace{0mm}
\end{figure*}

Rather than predicting the next token in a modality-agnostic way, we make predictions in a modality-aligned way. Namely, for each input token we predict the next token of the \emph{same} modality. Please see Figure~\ref{fig:arch} for diagrams.

\subsection{Joint training}

We have two options for training on collections that contain diverse trajectories in terms of noise levels or modality subsets. We can either train jointly with all data at once, including complete and incomplete trajectories. Alternatively, we can first pre-train on noisy and incomplete trajectories. This can be viewed as providing a good initialization for then training on complete trajectories. We find that both approaches work comparably in our setting and opt for joint training in the majority of the experiments for simplicity.

\subsection{Model architecture}

Our model is a vanilla transformer~\cite{Vaswani2017}. Given the trajectories from either complete or incomplete data, we first tokenize the trajectories into tokens. We learn separate linear projection layers for each modality but shared across time. To encode the temporal information we use positional embeddings. Let's assume $o_i \in \mathcal{R}^{m}$ and $a_i \in \mathcal{R}^{n}$, then:
\begin{align}
    t_i &= \texttt{concat}(o_i, a_i), \\
    h^0_i &= W t_i,
\end{align}
where $W \in \mathcal{R}^{d \times (m + n)}$ is a linear projection layer to project concatenated observation and action modalities to $d$ dimensional embedding vector. The superscript $0$ indicates the embedding at $0$-th layer, \ie, the input layer. When action is unavailable, we use a mask token $\texttt{[M]} \in \mathcal{R}^{n}$ to replace $a_i$, and $\texttt{[M]}$ is initialized as a random vector and learned end-to-end with the whole model.
The model takes the sequence of embedding vectors $H_0 = \{ h^0_1,  h^0_2, ..., h^0_t\}$ as input.

The transformer architecture contains $L$ layers, each consisting of a multi-head self-attention module and an MLP module. Assume the output of the layer $l$ is $H_l$, then the layer $l+1$ output is computed as follows:
\begin{align}
    \Tilde{H_l} &= \texttt{LayerNorm} (H_l) \\
    \Tilde{H_l} &= \Tilde{H_l} + MHSA(\Tilde{H_l}) \\
    H_{l+1}     &= \Tilde{H_l} + MLP(\Tilde{H_l}) 
\end{align}

Here, the multi-head self-attention has causal masking, where the token only attends to itself and the past tokens. Once the tokens are processed through all the layers, we project the embedding to predicted states and actions, by learning a linear projection layer $\widehat{W} \in \mathcal{R}^{(m + n) \times d}$:
\begin{align}
    \widehat{t}_{i+1} &= \widehat{W} h^L_i \\
    \widehat{o}_{i+1} &= (\widehat{t}_{i+1})_{0:m} \\
    \widehat{a}_{i+1} &= (\widehat{t}_{i+1})_{m:(m+n)}
\end{align}

Then we train the transformer with the objective in \eqref{eq:loss}. In the cases where the token is masked, we do not apply any losses. We train our transformer with both types of data, as shown in Figure~\ref{fig:arch}. This allows us to use various sources of data, thus enabling scaling in terms of data.

\begin{figure*}[t]\centering\vspace{0mm}
\includegraphics[width=1.0\linewidth]{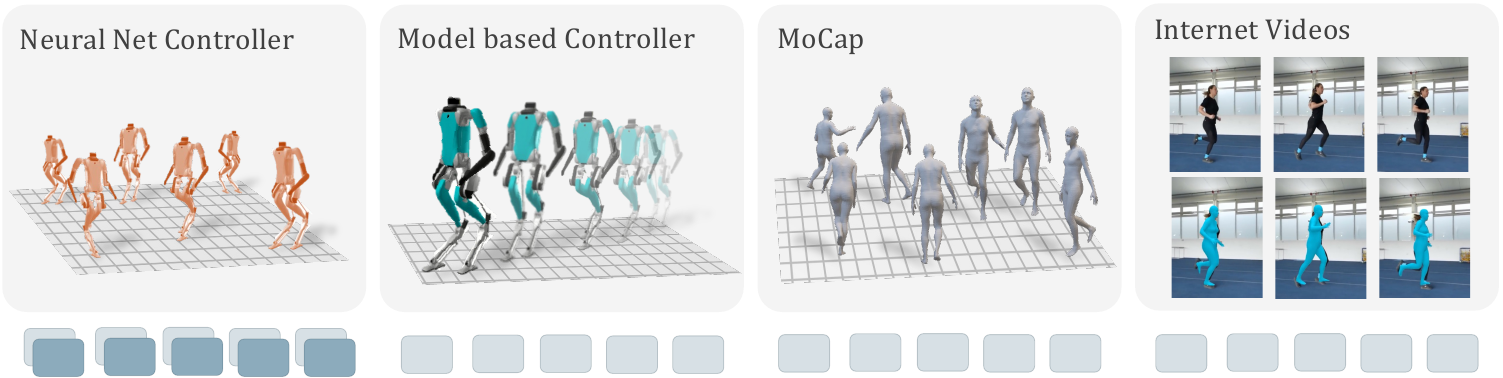}\vspace{-0mm}
\caption{\textbf{Training dataset.} To train our model, we construct a dataset of trajectories coming from four different sources. \emph{(i) neural network policy:} provides trajectories with complete observations and actions. \emph{(ii) model-based controller:} produces trajectories without actions. \emph{(iii) motion capture of humans:} does not contain actions and is approximately retargeted onto the robot. \emph{(iv) internet videos of humans:} noisy human poses are first reconstructed via 3D reconstruction and then approximately retargeted onto the robot.}
\label{fig:data}\vspace{0mm}
\end{figure*}

\subsection{Model inference}

At inference time, our transformer model will always have access to observation-action pairs. In this setting, we apply our transformer model autoregressively for each observation-action pair token. By conditioning on past observations and actions, we predict the next actions (or observation-action pair) and execute the action. Then we take the observations from the robot and discard the predicted observations. We use the observed observation and predicted action as the next set of tokens and concatenate them with past pairs to predict the next observation-action pair.

\section{Dataset}\label{sec:dataset}

Our approach motivates building a dataset of trajectories for training our model. Our dataset includes trajectories from different sources: (i) neural network policies, (ii) model-based controllers, (iii) human motion capture, and (iv) human videos from YouTube. An illustration of different data sources is shown in Figure~\ref{fig:data}. We describe each in turn next.

\subsection{Neural network trajectories}

As the first source of training trajectories, we use a neural network policy trained with large-scale reinforcement learning~\cite{RealHumanoid2023}. Specifically, this policy was trained with billions of samples from thousands of randomized environments in Isaac Gym~\cite{Makoviychuk2021}. We run this policy in the Agility Robotics' simulator and collect 10k trajectories of 10s each on flat ground, without domain randomization. Each trajectory is conditioned on a velocity command sampled from a clipped normal distribution as follows: linear velocity forward $[0.0, 1.0]$ m/s, linear velocity sideways $[-0.5, 0.5]$ m/s, and turning angular velocity $[-0.5, 0.5]$ rad/s.

Since we have access to the data generation policies, we are able to record complete observations as well as the exact actions that the model predicted. We use this set as our source of complete sensorimotor trajectories that have complete observations as well as ground truth actions.

\subsection{Model-based trajectories}

As the second source of trajectories, we use the model-based controller developed by Agility Robotics. It is the controller that is deployed on the Digit humanoid robot and available in the Agility Robotics' simulator as well. We collect two sets of 10k trajectories of walking on a flat ground of 10s each. In both cases, we sample the velocity commands as follows: linear velocity forward $[-1.0, 1.0]$ m/s, linear velocity sideways $[-1.0, 1.0]$ m/s, and turning angular velocity $[-1.0, 1.0]$ rad/s. We use the default model-based configurations for one set and randomize the leg length, step clearance, and bounciness of the floor for the other.

As this controller outputs joint torques, which are not consistent with our joint position action space. We only record the observations without the actions. This data serves as a source of trajectories with reasonably good observations from the same morphology but without the actions.

\subsection{Human motion capture trajectories} 

As the next source of trajectories, we use the motion capture (MoCap) recordings of humans from the KIT dataset~\cite{Plappert2016} distributed via the AMASS repository~\cite{AMASS:ICCV:2019}. This data was recorded using optical marker-based tracking in a laboratory setting. The dataset consists of $\app$4k trajectories. We use a subset of $\app$1k standing, walking, and running trajectories.

In addition to not containing the ground truth actions, the MoCap trajectories come with an additional challenge: different morphology. Namely, MoCap trajectories capture \emph{human} keypoint positions in 3D. In order to use these trajectories for training a robot, we solve an inverse kinematics problem to find the corresponding robot poses.

We formulate an inverse kinematics optimization problem:
\begin{subequations}
\label{eq:ko}
\begin{align}
    \min_{\substack{\mathbf{q}[t], \mathbf{\dot{q}}[t]}} ~ & \sum_{t=1}^{N} \varphi^{\text{traj}}[t] + \varphi^{\text{reg}}[t] \label{eq:ko-cost}\\
    \text{s.t.} ~
    & \mathbf{q}[t+1] = \mathbf{q}[t] + \frac{\mathbf{\dot{q}}[t+1] + \mathbf{\dot{q}}[t]}{2} dt, \label{eq:ko-posture-integration} \\
    & \mathbf{q} \in \mathcal{Q}, \mathbf{\dot{q}} \in \mathcal{V} \label{eq:ko-posture-constraint}
\end{align}
\end{subequations}
\noindent
where $\mathbf{q}$ is the robot state in the generalized coordinates, and $N$ and $dt$ are the optimization horizon and sampling time. The optimization variables include $\mathbf{q}$, $\mathbf{\dot{q}}$. 
For constraints, \eqref{eq:ko-posture-integration} is the Euler integration of posture $\mathbf{q}$, \eqref{eq:ko-posture-constraint} constrains the range of $\mathbf{q}$ and $\mathbf{\dot{q}}$ to their admissible sets $\mathcal{Q}$ and $\mathcal{V}$. 
In the cost function, $\varphi^{\text{traj}}$ tracks keypoint locations from human trajectories, and $\varphi^{\text{reg}}$ represents the regularization costs, such as joint velocity minimization and smoothness.

\subsection{Trajectories from YouTube videos}

Internet videos of people doing various activities are potentially a vat source of data for learning human locomotion. However, the raw pixels have no information about the state and actions of the human. To recover this, we first we run a computer vision tracking algorithm PHALP~\cite{rajasegaran2022tracking} to extract human trajectories in 3D. This provides an estimate of the 3D joints of the human body SMPL~\cite{loper2023smpl} parameters and a noisy estimate of the human joints in the world coordinates. We use the human body joint positions to retarget the motion to the humanoid robot using the inverse kinematics optimization described above. Once we retarget the motion from the Internet videos to humanoid trajectories, we filter the trajectories with the low optimization cost. Note that the scale of this data comes with the cost of being noisy.

\begin{figure}[t]\centering\vspace{0mm}
\includegraphics[width=1.0\linewidth]{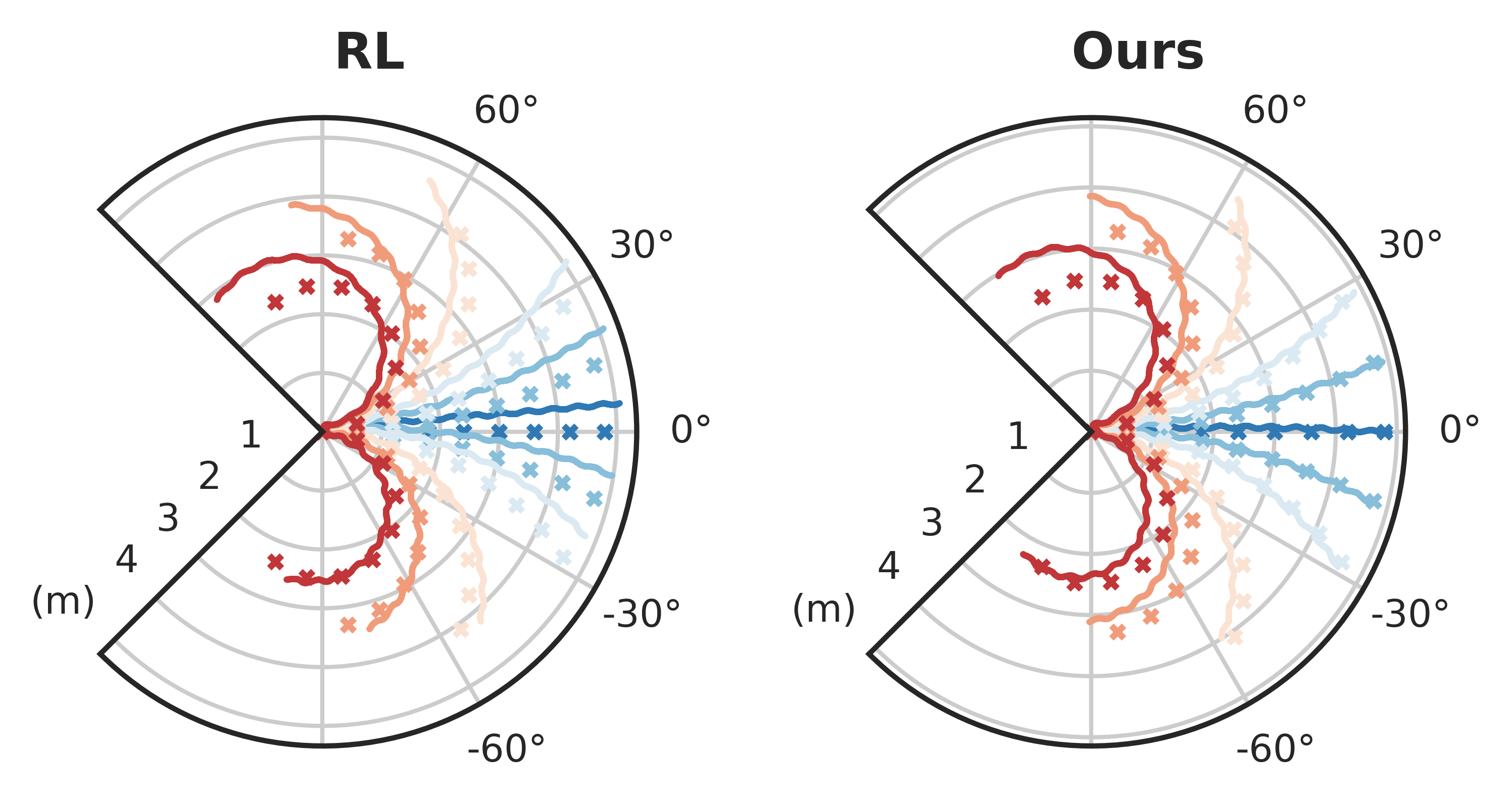}\vspace{-3mm}
\caption{\textbf{Comparison to state of the art, trajectory adherence.} The robot is commanded to walk starting from the origin with a fixed heading command of $0.5$ m/s and varying yaw commands in $[-0.4, 0.4]$ rad/s. We plot the desired (dotted) and actual (solid) trajectories for our policy and a reinforcement-learning trained policy (RL).}
\label{fig:trajectory}\vspace{0mm}
\end{figure}

\section{Experiments}

We evaluate our approach on the challenging task of humanoid locomotion. We perform outdoor experiments on real hardware and systematic evaluations in simulation.

\subsection{Experimental setup}

\textbf{Robot platform.} Digit is a humanoid robot platform developed by Agility Robotics. It is a full-sized humanoid that is 1.6m tall and weighs 45 kilograms. It has 30 degrees of freedom of which 20 are actuated. Due to its high dimensionality and four-bar linkage structure, it is challenging to optimize fast which makes it particularly interesting for learning approaches that can learn efficiently from trajectory collections like ours.

\textbf{Model.} Our model has a hidden size of $192$ dimensions, with 4 layers of self-attention layers and MLP layers. Each self-attention has 4 heads. We use LayerNorm before each attention layer and ReLU activation after the MLP layer. We use a BatchNorm layer to process the input before the transformer model. When predicting a token at time $k$, to keep the context length at a reasonable size, we only keep the past 16 steps in input. In Section \ref{sec:scaling}, we show the model is able to scale up to more parameters and longer context length and achieve higher performance.

\subsection{Real-world deployment}

We begin by reporting the results of deploying our policy in the real world. Specifically, we evaluate deploying our robot at various locations in San Francisco over the course of one week. Please see Figure~\ref{fig:sf} for examples and \href{humanoid-next-token-prediction.github.io}{project page} for videos. We find that our policy is able to walk over a variety of surfaces including walkways, concrete, asphalt, tiled plazas, and dirt roads. Note that the deployment in a large city environment, like San Francisco, is considerably more challenging than in constrained environments. The city environment is much more crowded, less patient, and not forgiving. This makes the error tolerance low and requires a policy that works consistently well.

\begin{figure}[t]\centering\vspace{2mm}
\includegraphics[width=\linewidth]{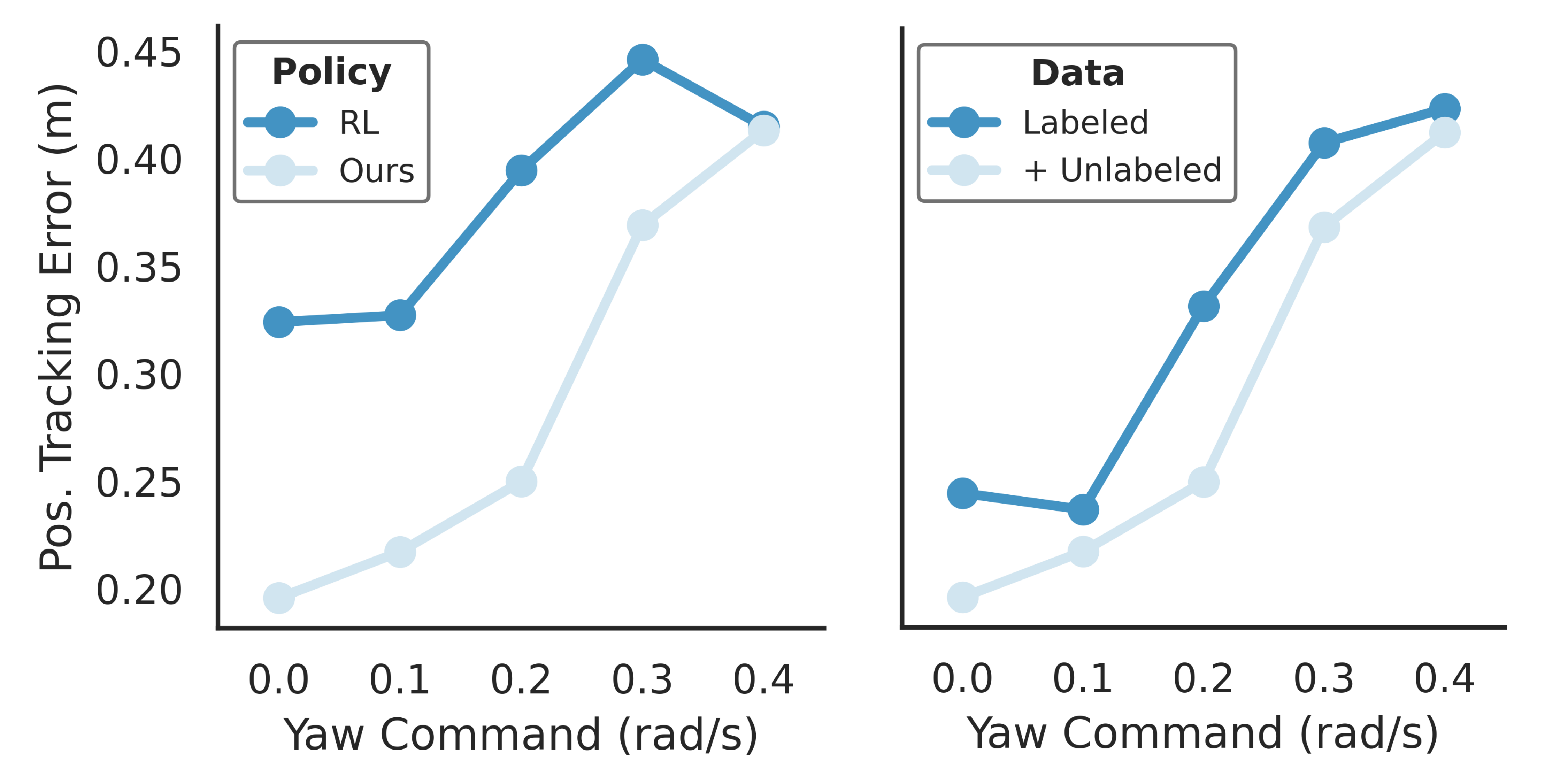}\vspace{-2mm}
\caption{\textbf{Tracking error comparisons.} We measure the tracking error of our policy against a state-of-the-art benchmark (left), as well as the improvement produced by complementing action-labeled RL trajectories with action-free trajectories (right).}
\label{fig:tracking}\vspace{0mm}
\end{figure}

\subsection{Evaluation Metrics} \label{sec:metrics}

We evaluate locomotion policies with two metrics: \textit{tracking error} and \textit{prediction error}. Tracking error measures how accurately the robot follows a specific locomotion command. The prediction error is the next token prediction loss measured on a separate set of validation data. We introduce two metrics with details as follows and show that two metrics can consistently predict locomotion performance.

\textbf{Tracking error.} In all experiments, the robot starts from rest in a simulated environment and is issued a constant natural walking command consisting of a desired heading velocity sampled in $[0.35, 0.70]$ m/s, angular velocity sampled in $[-0.4,0.4]$ rad/s, and zero lateral velocity. We compute $\mathbf{x^{*}}(t)$, the ideal robot base position trajectory that fully satisfies the velocity command $\mathbf{v^{*}}(t)$ at all time steps. To measure the accuracy of command tracking, we define the position tracking error as $\frac{1}{T} \sum_{t=0}^{T}{\lVert \mathbf{x}(t) - \mathbf{x^{*}}(t) \rVert}$. We use the MuJoCo simulator~\cite{Todorov2012} for evaluations, and all trajectories last for a duration of 10 seconds.

\textbf{Prediction error.} Since the model is trained with the next token prediction, we test the prediction error on a set of validation data that is separated from training data and contains state-action trajectories collected from the RL policy. This is similar to the language modeling evaluation for large language models~\cite{hendrycks2020measuring}. We test both state and action prediction errors and add them together as the final error metric.

\subsection{Comparison to the state of the art}

\textbf{Trajectory Adherence.} We compare our policy to a neuralfig:tracking network controller trained with reinforcement learning (RL)~\cite{RealHumanoid2023}. Figure \ref{fig:trajectory} presents a visual comparison of the trajectory adherence of our controller against these state-of-the-art baselines. Starting with a robot at the origin, we plot the actual trajectory of the robot with eleven different yaw commands selected from $\{0.00, \pm0.05, \pm0.10, \pm0.20, \pm0.30, \pm0.40\}$ rad/s. For each policy, we jointly plot the desired and actual path traced by the robot base. Our model exhibits superior tracking to the RL controller at all turning speeds, and has near-perfect tracking for straight-line walking.  

\textbf{Quantitative Evaluation.} In Figure \ref{fig:tracking}, left, we repeat the above comparison to the RL controller $(N=245)$, with the full range of heading and yaw velocities mentioned in Section~\ref{sec:metrics}. We plot the mean position tracking error, binned by the commanded angular yaw. While both models have lower tracking errors at lower yaw, ours consistently outperforms the baseline RL policy. This is an interesting result, since our model was trained on next token prediction on trajectories produced by this very policy.

\begin{figure}[t]
\centering
\includegraphics[width=0.8\linewidth]{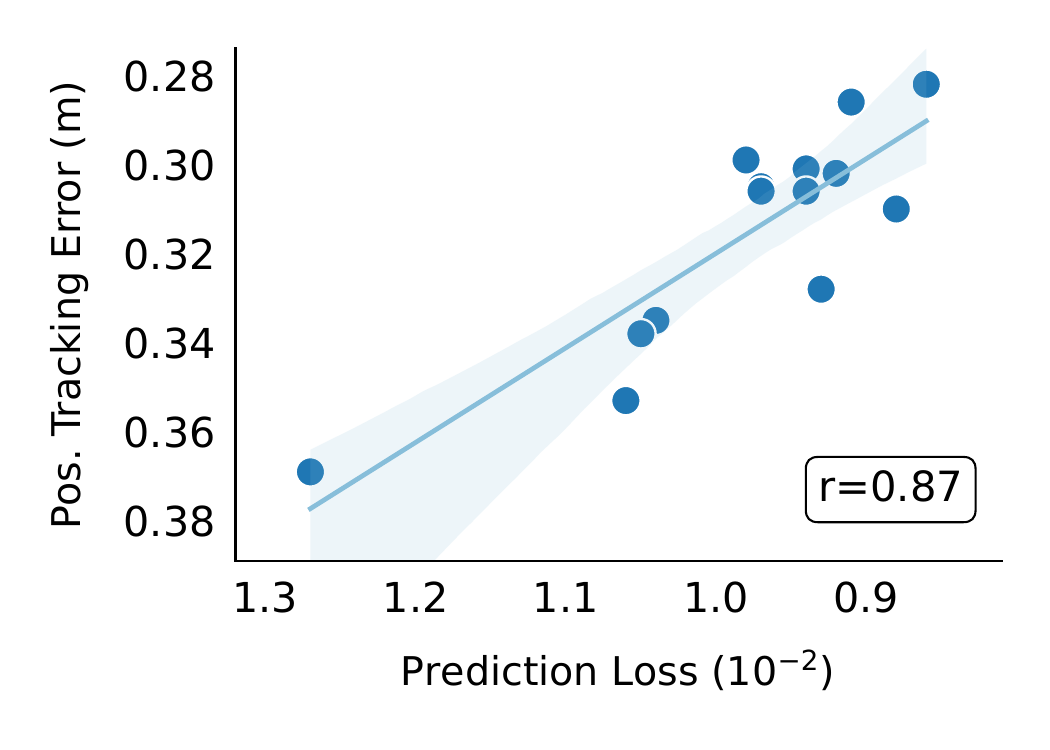}
\caption{\textbf{Prediction error correlates with performance.} We plot the tracking error and prediction error for 14 models. The prediction error linearly correlates with task tracking error with $r=0.87$, which means lower prediction loss likely indicates more accurate command following.}
\vspace{0mm}
\label{fig:metrics_correlation}
\end{figure}

\subsection{Prediction error correlates with performance}

We collect 14 models trained with different training recipes, model architectures, data size and types, and test tracking error and prediction error for each one of them. We plot the tracking and prediction errors of all the models into a single scatter plot, as shown in Figure \ref{fig:metrics_correlation}. We can see that tracking and prediction error are highly correlated with Pearson coefficient $r = 0.87$, which means models with lower prediction error on the validation set likely follow different commands with higher accuracy. This suggests that the prediction error is predictive task performance.

\begin{figure}[t]\centering\vspace{0mm}
\includegraphics[width=1.0\linewidth]{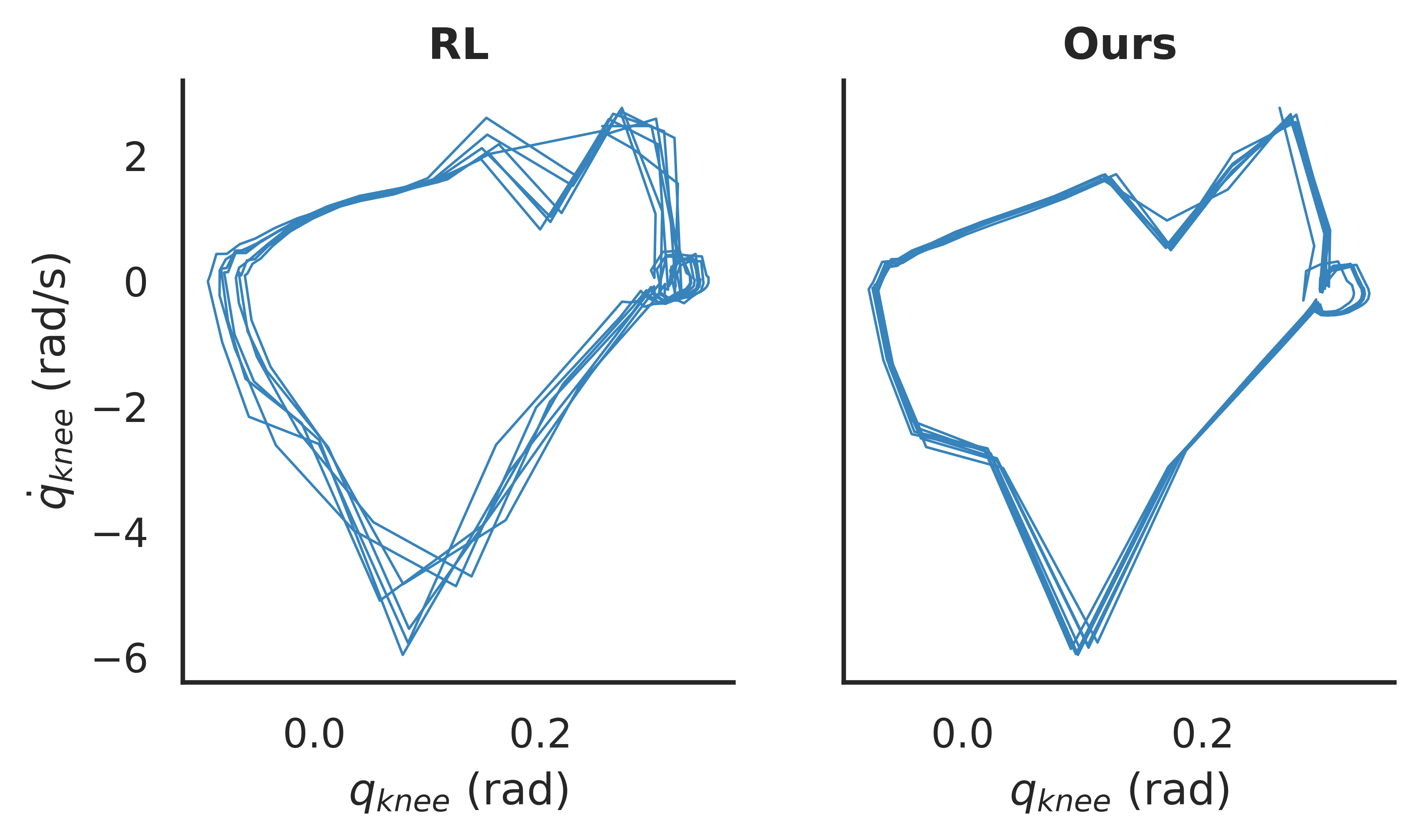}\vspace{-0mm}
\caption{\textbf{Gait quality.} We command the robot with a heading velocity of $0.5$ m/s and plot the resulting phase portrait of the left knee joint. Compared to the RL policy, our policy features fewer irregularities and a smoother, cyclic gait.}
\label{fig:gait_quality}\vspace{-2mm}
\end{figure}

\begin{figure}[b]\centering\vspace{0mm}
\includegraphics[width=0.8\linewidth]{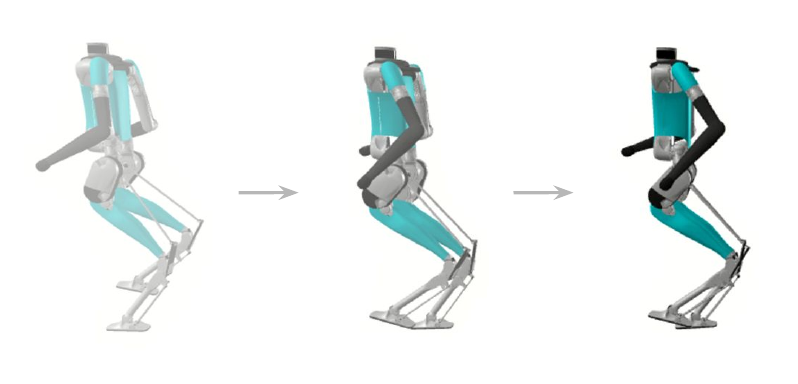}\vspace{-0mm}
\caption{\textbf{Unseen commands.} Our policy is able to follow backward commands at test time, unseen during training.}
\label{fig:unseen}\vspace{0mm}
\end{figure}

\begin{figure*}[t]\centering\vspace{0mm}
\includegraphics[width=0.9\linewidth]{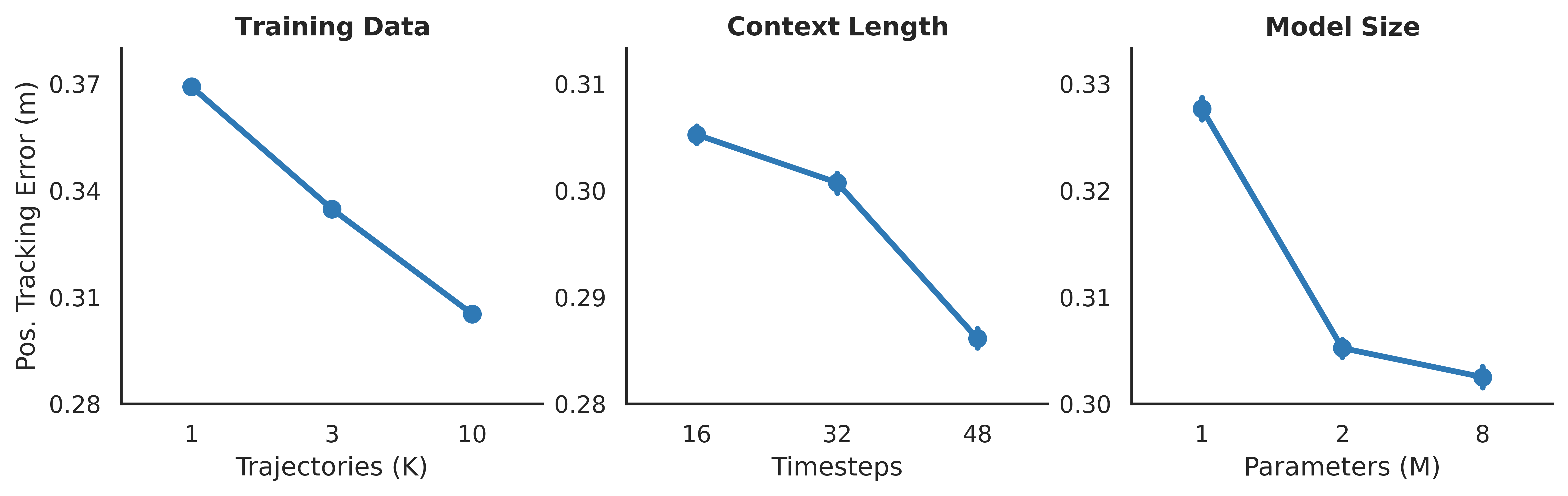}\vspace{-0mm}
\caption{\textbf{Scaling studies.} We find that our approach scales with the number of trajectories in the training dataset (left), context length (middle), and larger models (right).}
\vspace{-0.5em}
\label{fig:scaling}\vspace{0mm}
\end{figure*}

\subsection{Gait quality}

In humanoid locomotion, the smoothness in the robot's gait is contingent on the rhythmic functioning of its actuated knee joints. One way to measure this is a phase portrait, which is a parametric plot of a joint's generalized position and velocity over time. Patterns in the plot can reveal information about the type of movement the joint is undergoing. For example, a cyclic pattern may indicate repetitive motion, while irregular patterns might suggest complex or varied movements, such as stumbling. In Figure \ref{fig:gait_quality}, we command the robot to walk forward at $0.5$ m/s, and plot the associated phase portrait of its left knee joint. Notice that our policy retains the overall shape of the RL policy while having fewer aberrations. This supports our qualitative assessment of the more regularized behavior seen on our policy.

\subsection{Generalization to unseen commands}

We find that our policy also extrapolates new skills such as walking backward, which was \emph{not} included in the action-labeled training data. As Figure \ref{fig:unseen} illustrates, by prompting our controller with negative values for the heading command, we find that the robot naturally performs backward walking at speeds up to 0.5 m/s without falling.

\subsection{Training with action-free data}

One of the benefits of our approach is that it can be applied to trajectories from diverse sources, including missing information like actions in the case of human videos from YouTube. In Figure~\ref{fig:tracking}, right, we compare the performance of training only with complete trajectories to joint training on both complete and incomplete trajectories. We observe that including incomplete trajectories consistently leads to better performance. This is a promising signal for scaling our approach to a large collection of diverse trajectories.

\subsection{Scaling studies}\label{sec:scaling}

\textbf{Training data.} In Figure~\ref{fig:scaling}, left, we study the scaling of our model's performance by increasing the size of the training dataset. We find that training on more trajectories reduces position tracking error, which is a positive signal for increased performance when training on larger datasets.

\textbf{Context length.} We study the effect of increasing the number of tokens used in the context window of the transformer policy, varying it between 16, 32, and 48 steps in Figure \ref{fig:scaling} middle. Larger context windows produce better policies, which suggests that our generative policy performs a form of in-context adaptation that improves with scale.

\textbf{Model size.} We compare models with increasing number of parameters (1M, 2M, 8M) by varying the embedding dimension (144, 192, 384), number of attention heads (3, 4, 12), and number of transformer blocks (4, 4, 6) respectively. Tracking error monotonically decreases with model size.

\begin{table*}[htbp] 
\centering 

\vspace{2em}

\begin{subtable}{0.475\textwidth}
    \centering
    \begin{small}
    \begin{tabular}{p{0.27\textwidth}cc}
        \toprule
         & Track Err. & Pred. Err. \\
        \midrule
        Concat & 0.310 & \textbf{0.88} \\
        Separate & \textbf{0.299} & 0.98 \\
        \bottomrule
    \end{tabular}
    \end{small}
    \vspace{0.5em}
    \caption{\textbf{Concatenated \vs separate tokens} for states and action. Two modeling designs have comparable performance while concatenating state and action gives shorter input length and faster inference.}
    \label{tab:ablation_concat_separate}
\end{subtable}%
\hfill
\begin{subtable}{0.475\textwidth}
    \centering
    \begin{small}
    \begin{tabular}{p{0.27\textwidth}cc}
        \toprule
         & Track Err. & Pred. Err. \\
        \midrule
        Align & \textbf{0.299} & \textbf{0.98} \\
        Non-align & 0.338 & 1.05 \\
        \bottomrule
    \end{tabular}
    \end{small}
    \vspace{0.5em}
    \caption{\textbf{Alignment \vs non-alignment} of states or actions for next token prediction. Prediction with aligned modality performs better on bothe tracking error and next token prediction error.}
    \label{tab:ablation_alignment}
\end{subtable}

\vspace{2em} 

\begin{subtable}{0.475\textwidth}
    \centering
    \begin{small}
    \begin{tabular}{p{0.27\textwidth}cc}
        \toprule
         & Track Err. & Pred. Err. \\
        \midrule
        Joint training & \textbf{0.310} & 0.88 \\
        Staged training & 0.311 & - \\
        \bottomrule
    \end{tabular}
    \end{small}
    \vspace{0.5em}
    \caption{\textbf{Joint \vs staged training} on data with and without actions. Staged training which pre-trains on state prediction and finetunes on action prediction has similar performance as joint training.}
    \label{tab:ablation_joint_staged}
\end{subtable}%
\hfill
\begin{subtable}{0.475\textwidth}
    \centering
    \begin{small}
    \begin{tabular}{p{0.27\textwidth}cc}
        \toprule
         & Track Err. & Pred. Err. \\
        \midrule
        State-action & \textbf{0.305} & 0.97 \\
        Action-only & 0.335 & - \\
        \bottomrule
    \end{tabular}
    \end{small}
    \vspace{0.5em}
    \caption{\textbf{State-action \vs action-only prediction}. Predicting both states and actions leads to lower tracking error than only predicting action as in vanilla behavior cloning.}
    \label{tab:ablation_state_action_pred}
\end{subtable}

\vspace{1em}
\caption{\textbf{Ablations on different design choices in modeling and training}. For each ablation we compare the average tracking error on a set of commands, as well as the next token prediction error on the test set.  For a fair comparison, we do not report next token prediction error for models that only predict actions.}
\label{tab:ablations}
\end{table*}

\subsection{Ablation studies}

\textbf{Concatenated \vs separate tokens.} For the input of transformer, we can either concatenate observation and action at each step into a single token, or embed them into two separate tokens. We compare these two choices in Table \ref{tab:ablation_concat_separate}. We can see that concatenation has lower prediction error while separating tokens has lower tracking error. Overall these two perform comparably while using separate tokens doubles the input length and introduces computation overhead. 

\textbf{Modality-aligned \vs non-aligned prediction.} When we use separate tokens for observation and actions as input, we can either predict $\widehat{o}_{i+1}$ from $o_i$ and $\widehat{a}_{i+1}$ from $a_i$, which aligns modality between prediction and input, or we can predict $\widehat{o}_{i+1}$ from $a_i$ and $\widehat{a}_{i+1}$ from $o_{i+1}$, which does not have alignment. From Table \ref{tab:ablation_alignment}, we can see that modality alignment has clearly better performance than no alignment. We suspect this is because, at $t$-th step during inference, when predicting action of $(t+1)$-th step, since there is no alignment, we need to first predict $\widehat{o}_{i+1}$ and use this prediction as input to predict $\widehat{a}_{i+1}$. If the predicted $\widehat{o}_{i+1}$ is not accurate compared to real $o_{i+1}$ (which is used to predict $\widehat{a}_{i+1}$ during training), there will be a discrepancy between test and training data which will cause error in action prediction.

\textbf{Joint training \vs staged training.} Given both complete data with action and incomplete data without action, we can either jointly train on both data as described in Section \ref{sec:approach}, or we can first pre-train the model on all the data with state prediction only, then fine-tune the model on complete data with action prediction. We compare these two approaches in Table \ref{tab:ablation_joint_staged}. We observe no significant difference between these two, which indicates that pre-training on state prediction then fine-tuning on action prediction also gives a reasonable locomotion policy.

\textbf{State-action prediction \vs action-only prediction.} We compare the performance of our policy when trained with only predicting actions, versus when trained with predicting both states and actions. The results in Table~\ref{tab:ablation_state_action_pred} show that the state-action prediction improves model performance on trajectory tracking. We hypothesize that the additional learning signal enables the model to learn richer representations of the world that are beneficial for the locomotion task.

\section{Discussion}

We present a self-supervised approach for real-world humanoid locomotion. Our model is trained on a collection of sensorimotor trajectories, which come from prior neural network policies, model-based controllers, human motion capture, and YouTube videos of humans. We show that our model enables a full-sized humanoid to walk in the real-world zero-shot. These findings suggest a promising path toward learning challenging real-world robot control tasks by generative modeling of large collections of trajectories.

\section*{Acknowledgements}

This work was supported in part by DARPA Machine Common Sense program, ONR MURI program (N00014-21-1-2801), NVIDIA, Hong Kong Centre for Logistics Robotics, The AI Institute, and BAIR’s industrial alliance programs. We thank Saner Cakir and Vikas Ummadisetty for help with the inverse kinematics simulation experiments.

\bibliography{references}
\bibliographystyle{icml2024}

\end{document}